\documentclass[runningheads]{llncs}
\usepackage{multirow} 
\usepackage{amsmath} 
\usepackage{amsfonts}
\usepackage{amssymb} 
\usepackage{booktabs}
\usepackage{bbding}
\usepackage{pifont}
\usepackage[table]{xcolor}
\usepackage{stfloats}
\usepackage{cite}
\usepackage[T1]{fontenc}
\usepackage{wrapfig}  
\usepackage{array}    
\usepackage{tikz}
\usepackage{pgfplots}
\usepgfplotslibrary{fillbetween} 

\usepackage{graphicx}
\usepackage{graphicx,verbatim}
\usepackage{xcolor}
\usepackage{hyperref}

\begin{document}
\title{UltraAD: Fine-Grained Ultrasound Anomaly \\Classification via Few-Shot CLIP Adaptation
}

%

\author{Yue Zhou\inst{1, 2} \and Yuan Bi\inst{1, 2} \and Wenjuan Tong\inst{3} \and \\ Wei Wang\inst{3} \and Nassir Navab\inst{1} \and Zhongliang Jiang\inst{1}}  

\institute{Computer Aided Medical Procedures (CAMP),\\ TU Munich, Germany \\
\email{zl.jiang@tum.com} \\
\and Munich Center for Machine Learning (MCML), Munich, Germany 
\and The First Affiliated Hospital of Sun Yat-Sen University,
Guangzhou, China}

\authorrunning{Yue Zhou et al.}
\titlerunning{Fine-Grained Ultrasound Anomaly Classification}
\maketitle              
\begin{abstract}
Precise anomaly detection in medical images is critical for clinical decision-making. While recent unsupervised or semi-supervised anomaly detection methods trained on large-scale normal data show promising results, they lack fine-grained differentiation, such as benign vs. malignant tumors. Additionally, ultrasound (US) imaging is highly sensitive to devices and acquisition parameter variations, creating significant domain gaps in the resulting US images. To address these challenges, we propose UltraAD, a vision-language model (VLM)-based approach that leverages few-shot US examples for generalized anomaly localization and fine-grained classification. To enhance localization performance, the image-level token of query visual prototypes is first fused with learnable text embeddings. This image-informed prompt feature is then further integrated with patch-level tokens, refining local representations for improved accuracy. For fine-grained classification, a memory bank is constructed from few-shot image samples and corresponding text descriptions that capture anatomical and abnormality-specific features. During training, the stored text embeddings remain frozen, while image features are adapted to better align with medical data. UltraAD has been extensively evaluated on three breast US datasets, outperforming state-of-the-art methods in both lesion localization and fine-grained medical classification. Project page: \href{https://karolinezhy.github.io/UltraAD/}{https://karolinezhy.github.io/UltraAD/}


\end{abstract}

\section{Introduction}
\noindent
Medical ultrasound (US) is a widely used imaging modality for examining internal organs, such as the breast and thyroid, due to its real-time capability, non-radiative nature, and accessibility. In remote or low-income regions, it is often the only available diagnostic tool. However, ultrasound images frequently suffer from low quality and significant domain variations due to differences in imaging devices and acquisition parameters~\cite{jiang2023robotic, bi2024machine}, which pose challenges for ultrasound image understanding~\cite{li2025semanticscenegraphultrasound}. To address these challenges, a generalized anomaly detection (AD) algorithm that adapts across diverse anatomical structures and imaging domains is highly demanded for supporting clinical decision-making, particularly for junior clinicians with limited clinical experience.

\par
While supervised deep learning has achieved phenomenal success in medical image segmentation \cite{huang2025vibnet, lee2020channel, zhuang2019rdau, tang2023cmu, huang2023motion, jiang2024intelligent}, these methods require large labeled training data, which is costly in the US imaging field. To overcome this limitation, unsupervised and self-supervised approaches leveraging autoencoders and generative models have been developed using only unlabeled data \cite{bi2024synomaly, bercea2025evaluating}. Despite success in MRI anomaly detection, these methods cannot be directly adopted to US images because of the noisy nature and limited imaging field of view.

\par
To improve generalization across unseen domains or objects, recent advancements in vision-language models (VLMs) have gained increasing attention. A pioneering study, WinCLIP \cite{jeong2023winclip}, employed pre-trained CLIP \cite{radford2021learning} with binary textual prompts for anomaly detection. Subsequent methods, including VAND \cite{chen2023april}, AnomalyCLIP \cite{zhou2023anomalyclip}, AdaClip \cite{cao2024adaclip}, and VCP-CLIP \cite{qu2024vcp}, have further integrated prompt learning and adapter mechanisms to better align natural image representations with specific application domains. MediCLIP \cite{zhang2024mediclipadaptingclipfewshot} adapts the pretrained CLIP model for anomaly detection in medical images by leveraging few-shot normal images with synthetic anomalies. The improved generalization of these approaches is largely attributed to the universality of text prompts across diverse datasets. However, these methods focus on binary anomaly detection, neglecting the need for fine-grained classification, which is crucial for distinguishing lesion phases or benign and malignant tumors in medical applications.

\par
For few-shot classification tasks using VLMs, Radford~\emph{et al.} introduced linear probing, which optimizes a classifier on top of frozen vision encoders \cite{radford2021learning}. Building on this, LP++ \cite{huang2024lp++} has demonstrated promising results in the natural image domain. To better align language and image features while preserving pretrained representations, prompt learning methods incorporate learnable tokens \cite{zhou2022conditional, yao2023visual, zhou2022learning}. To enhance few-shot classification, CLIP-Adapter \cite{gao2024clip} introduced lightweight non-linear projections via MLP layers. Alternatively, Tip-Adapter \cite{zhang2021tip} offers a training-free approach that achieves comparable classification performance through direct feature optimization. In the medical domain, few-shot adaptation has also demonstrated strong performance on surgical data, as evidenced by recent work \cite{chen2025textdrivenadaptationfoundationmodels, yuan2025learning}. Although promising results have been demonstrated on natural images, their extension to US imaging remains unexplored. Furthermore, due to the specific demands of clinical diagnosis, few-shot anomaly detection with fine-grained anomaly classification is crucial yet remains an unmet need in the research community.

\par
In this study, we introduce UltraAD, a CLIP-based framework for lesion localization and fine-grained anomaly classification through few-shot adaptation. Inspired by multi-task learning in medical image segmentation \cite{you2022class, jiang2024class}, UltraAD unifies pixel-wise anomaly localization and image-level classification to boost performance across tasks. To address the domain gap between natural and US images, a small set of US samples (4/8/16 shots in this study) is needed to adapt the CLIP model. To the best of our knowledge, this is the first CLIP-based model tailored for meeting US diagnosis requirements, leveraging domain-specific pathological information to improve generalization across imaging variations. Notably, the few-shot adaptation is performed on a single public breast dataset, while validation is conducted on two unseen breast datasets acquired using different machines, patients, and probe types. This highlights the robustness and effectiveness of the proposed method in real scenarios. The key contributions are: (1) A simple yet effective few-shot adaptation approach that generalizes across unseen US images of the same anatomy, despite significant domain variations in machine types and acquisition parameters; (2) a vision-language fusion strategy, combining learnable prompt tokens and image embeddings at both global (image-level) and local (patch-wise) scales to enhance anomaly localization; (3) a feature memory bank built from a small set of paired US images and text descriptions, enabling enhanced anomaly classification.

\section{Methodology}
\begin{figure}[t]
\includegraphics[width=\textwidth]{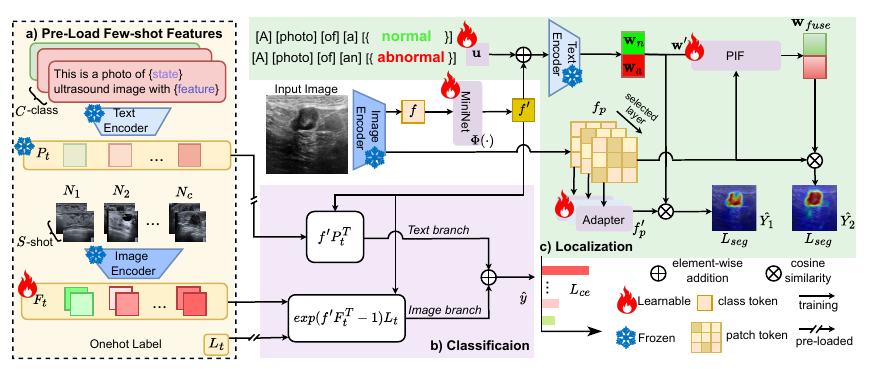}
\label{fig:pipeline}
\caption{\textbf{Proposed pipeline.} Via pretrained image and text encoders, text token embeddings $P_{t}$, image token embeddings $F_{t}$ and onehot label $L_t$ are preloaded from the S-shot C-class dataset. A mini-net projects image token embeddings $f$ to obtain $f'$. For anomaly classification, $F_{t}$, $P_{t}$ and $L_t$ are utilized to generate a classification score with $f'$ through feature similarity. For anomaly localization, two initial text prompts indicating normal and abnormal are added with the projected image token embedding $f'$. Two localization masks are produced by calculating the cosine similarity between an updated image token embedding $w'$ and the projected patch token $f'_p$, and a text-image fused text-embedding $w_{fuse}$ with the original patch token $f_p$. The final anomaly map prediction is computed as $\hat{Y}= \frac{1}{2}(\hat{Y}_1 + \hat{Y}_1).$}
\label{fig1}
\end{figure}

\noindent
\textbf{Problem Formulation} US domain gaps can arise from variations in machines and imaging parameters. We aim to develop a generalized model for pixel-wise anomaly localization and image-level classification indicating varying severity through few-shot CLIP adaptation. The $S$-shot training dataset from a specific domain is denoted as \( \mathcal{D}_{train} = \{ \mathbf{X}_i, \mathbf{Y}_i, y_i \}_{i=1}^{C \times S} \), where \( C \) represents the number of image classes. Here, \( \mathbf{X}_i \in \mathbb{R}^{H \times W \times 3} \) and \( \mathbf{Y}_i \in \mathbb{R}^{H \times W \times 1} \) are input image and its annotation mask, respectively.  
\( y_i \in \{0, 1, \dots, C-1\} \) is the class label. If \( y_i = 0 \), then \( \mathbf{Y}_i = \mathbf{0} \), indicating no anomaly; otherwise, \( y_i \) denotes a specific anomaly type, and \( \mathbf{Y}_i \) has pixels with a value of one, marking the anomaly region. The unseen test data is denoted as \( \mathcal{D}^{k}_{test} = \{ \mathbf{X}_i, \mathbf{Y}_i, y_i \}_{i=1}^{N_k} , \) which consists of images from the \( k \)-th domain (where \( k \in \{1,\ldots,K\} \) representing an unseen dataset). 

\subsection{CLIP for Anomaly Detection}
To adapt CLIP \cite{radford2021learning} for anomaly detection, two contrasting text prompts are used: A photo of a normal \textit{[CLS]} and A photo of an abnormal \textit{[CLS]}, where \textit{[CLS]} represents the specific image class category, e.g., breast US image. The text encoder \( \mathit{E_T}(\cdot) \) processes text prompts to generate the text embedding $\mathbf{w}_n \in \mathbb{R}^{D}$ for the normal class and $\mathbf{w}_a \in \mathbb{R}^{D}$ for the abnormal class. Given an input image \( X \), the image encoder \( \mathit{E_I}(\cdot) \) outputs a global class token \( f \in \mathbb{R}^{D} \) and local patch tokens \( f_p \in \mathbb{R}^{h \times w \times D} \), where $h$ and $w$ denote the spatial dimensions in feature space. The anomaly scores are obtained based on cosine similarity $\langle \cdot \rangle$ between text and visual embedding for both global image-level and local pixel-level, i.e.,\
\begin{equation}
\hat{y} = \frac{\exp(  \langle\mathbf{w}_a, f \rangle )}{  \exp(\langle\mathbf{w}_n, f \rangle) + \exp(\langle \mathbf{w}_a, f \rangle)},
\hat{Y} = \sigma(\frac{\exp(\langle\mathbf{w}_a, f_p \rangle )}{  \exp(\langle\mathbf{w}_n, f_p \rangle) + \exp(\langle \mathbf{w}_a, f_p \rangle)})
\end{equation}
where $\sigma(\cdot)$ is an interpolation function to upsample patch-level anomaly scores into the final predicted anomaly map $\hat{Y} \in \mathbb{R}^{H \times W}$.

\textbf{Image-Aware Prompting Module:} Inspired by \cite{zhou2022conditional}, to incorporate few-shot adaptation that improves generalization ability to unseen images, we proposed a simple unified template to generate both normal and abnormal text prompts based on input US images instead of using a pre-defined \textit{[CLS]} token. Specifically, the template replace the pre-defined \textit{[CLS]} token by learnable tokens, i.e.
\[
P = [\text{A}][\text{photo}][\text{of}][\text{a/an}][\textit{state}][u + f']
\]
where \( u \in \mathbb{R}^{D} \) is global learnable token and \( f' \in \mathbb{R}^{D} \) is instance-specific token. To obtain the instance-specific token \( f' \), a MiniNet \( \Phi(\cdot) \) takes the global image class token \( f \) as input, thus adding global image information into the text token.

\textbf{Image Feature Adapter:}
To further improve anomaly localization performance, we utilize the lightweight linear adapters~\cite{chen2023april} to refine image features obtained from pretrained CLIP image encoder. The linear adapters are optimized during few-shot adaptation. The adapters project patch token features \( f_p \) from each encoder layer \( l\) to $f'_p$ for anomaly map computation. Finally, the final anomaly map \( \hat{Y} \) is obtained by averaging anomaly maps from all layers.

\subsection{Memory-Boosted Few-Shot Adaptation}
Previous modifications are aimed at enhancing CLIP's performance in anomaly localization in US imaging. However, it still lacks the capacity for fine-grained anomaly classification, which is crucial for accurate disease grading and severity assessment. To address this limitation, our method utilizes memory-based few-shot adaptation and thus can predict different anomaly types.

\par
A support dataset with $S$-shot $C$-class training samples for few-shot adaptation denoted as $\mathcal{I}$ with corresponding labels $\mathcal{L}$, as illustrated in the left yellow part of Fig.\ref{fig:pipeline}. The total number of training data is $N = S \times C$. To enhance adaptation in the medical domain, beyond class labels, a lesion-aware prompt template $\mathcal{P}$ is incorporated to provide domain-specific contextual information for each class. The template is structured as: [This] [is] [a] [\textcolor{blue}{\textit{type}}] [ultrasound] [image] [with] [\textcolor{blue}{\textit{pathological features}}]. For instance, in the case of a benign lesion, the prompt is instantiated as: \textit{"This is a benign ultrasound image with round shape and sharply demarcated margins."}
We employ the pretrained CLIP text encoder $\mathit{E_T}(\cdot)$ and image encoder $\mathit{E_I}(\cdot)$ to obtain both text tokens and image tokens that are pre-computed and stored in the memory, i.e.,
\begin{equation}
\mathbf{F}_{\text{t}} = \mathit{E_I}(\mathcal{I}) \in \mathbb{R}^{N\times D},~ 
\mathbf{P}_{\text{t}} = \mathit{E_T}(\mathcal{P}) \in \mathbb{R}^{C\times D},~
\mathbf{L}_{\text{t}} = \text{OneHot}(\mathcal{L})\in \mathbb{R}^{N\times C}
\end{equation}
The classification result is determined by the similarity between the text prompt and the projected image class token $f'$ using a lightweight network. Additionally, it considers the interaction between the few-shot image memory and the image class token, as proposed in \cite{zhang2021tip}. The classification score for a sample is computed as follows, also shown in pink in Fig.\ref{fig:pipeline}:
\begin{equation}
    \hat{y} = f' \textbf{P}_{\text{t}}^T + \exp(f'\mathbf{F}_{\text{t}}^T-1) L_{\text{t}}
\end{equation}

\subsection{Patch-Wise Image-Language Fusion}
To enhance the alignment between US-informed text embeddings~\cite{zhou2022conditional} and image patch features, a further feature fusion of patch-wise image embedding and prompts representation is carried out (see PIF block in green in Fig.~\ref{fig1}). To effectively leverage the few-shot samples, we utilize class-specific prompts from fewshot prompt embedding $\textbf{P}_t$ by ensembling them with normal and abnormal prototypes for anomaly detection. The normal prototype is defined as \( w_n' = w_n + \textbf{P}_t^0 \), while the abnormal prototype is given by \( w_a' = w_a + \frac{1}{C-1
} \sum_{i=1}^{C} \textbf{P}_t^i \), ensuring diversity in normal and abnormal prompts by incorporating US-aware information. Hence, we get $ w' = [wn', wa']$. To further refine feature fusion, we adopt an M-head cross-attention module that learns three projection matrices, \( W_Q, W_K, W_V \), to compute image-language fused text embedding:
\begin{equation}
Q = \textbf{w}' W_Q, ~K = f_p W_K, ~ V = f_p W_V, ~ \textbf{w}^{fuse} = \text{softmax} \left( QK^T \right)  V
\end{equation}

\subsection{Multitask Learning for Anomaly Classification and Localization} 
\label{sec:enhancing}
\par
The class token, which acts as an embedding containing global image information, can be used for image classification and integrated into the prompt for conditioned prompt learning. To achieve this, we combine anomaly classification and anomaly localization by using a shared, dynamically updated class token within a unified framework. This approach seamlessly integrates anomaly classification with anomaly localization within a cohesive learning pipeline. The initially extracted CLIP image feature, denoted as $f$, is projected via a MiniNet $\Phi(\cdot)$, resulting in $f'$. 
During training, UltraAD refines the pixel-level anomaly maps $\hat{Y}_1$ and $\hat{Y}_2$ using a combination of Dice loss \cite{milletari2016v} and Focal loss \cite{lin2017focal}, equally weighted, on auxiliary data (few-shot samples) to ensure accurate segmentation. Meanwhile, it refines classification performance by employing cross-entropy loss. The image feature memory $\textbf{F}_t$ will be refined during the training while text feature $\textbf{P}_t$ remains frozen. The network is trained end-to-end with a joint optimization of segmentation and classification losses. The shared feature $f'$ is jointly refined, which will be used to compute the classification result as well as the segmentation result.

\section{Experiments}
\begin{table}[t]
    \centering
  \caption{Summary of breast ultrasound datasets used in experiments}
    \label{tab:datset}
    \resizebox{0.98\columnwidth}{!}{%
    \begin{tabular}{@{}ccccccc@{}}
    \toprule
    Dataset & Total & Normal & Benign    & Malignant   & Ultrasound System                        & Year \\ \midrule
    BUS-UCLM & 683 & 419 & 174 & 90  & Siemens ACUSON S2000TM     & 2022-23 \\
    BUSI     & 780 & 133 & 437 & 210 & LOGIQ E9, LOGIQ E9 Agile & 2018      \\
    BUSZS   & 300 & 100 & 100 & 100 & Mindray, Toshiba, GE, Canon, PHILIPS, Esaote & {2023}                                 \\
    \bottomrule
    \end{tabular}%
    }
\end{table}
\subsection{Experimental Setup}
\noindent
\textbf{Training and Testing}
Tab.\ref{tab:datset} details the datasets used for training and testing. We employ a few-shot subset of BUS-UCLM \cite{vallez2025bus} for model adaptation and assess performance on a public breast dataset BUSI \cite{al2020dataset} and an in-house dataset collected using six different US machines with both convex and linear probes, denoted as BUSZS without additional adaption. 
For the few-shot setting, we select 4, 8, and 16 samples per class from the BUS-UCLM \cite{vallez2025bus}. For fair comparisons, all baseline methods use identical training shots, except WinCLIP \cite{jeong2023winclip}, which only uses a pre-trained CLIP model without any further training. Other anomaly detection methods \cite{cao2024adaclip, zhou2023anomalyclip, qu2024vcp, huang2024adapting} and few-shot CLIP adaptation approaches \cite{zhou2022conditional, gao2024clip, huang2024lp++} require training on the few-shot data. For consistency, we use the ``ViT-L/14@336px'' CLIP backbone across all methods.

\par
Following the evaluation methods used in anomaly detection and localization tasks \cite{jeong2023winclip, zhou2023anomalyclip, cao2024adaclip, qu2024vcp}, we utilize the AUROC score for classification and AUROC/AUPRC for localization. All the values in Tab.~\ref{tab:cls_results} and Tab.~\ref{tab:seg} shows the average metrics of three experiments using 3 seeds for few-shot sample selection to ensure fair evaluation and reduce random few-shot sample variations.

\subsection{Performance Analysis}

\begin{table}[]
\caption{Comparison of anomaly detection (left) and classification (right) performance
across benchmarks, evaluated using image-level AUROC score. The best values are
\textbf{bold}, and the second best is \underline{underlined}.}
\centering
\resizebox{\columnwidth}{!}{%
\begin{tabular}{@{}cccccccclcccccccc@{}}
\toprule
\multicolumn{1}{l}{{\color[HTML]{000000} }} &
  \multicolumn{7}{c}{{\color[HTML]{000000} Anomaly Detection (binary)}} &
  {\color[HTML]{000000} } &
  {\color[HTML]{000000} } &
  \multicolumn{7}{c}{{\color[HTML]{000000} Anomaly Detection (multi-class)}} \\ \midrule
{\color[HTML]{000000} \textbf{}} &
  \multicolumn{3}{c}{{\color[HTML]{000000} BUSI}} &
  {\color[HTML]{000000} } &
  \multicolumn{3}{c}{{\color[HTML]{000000} BUSZS}} &
  {\color[HTML]{000000} } &
  {\color[HTML]{000000} } &
  \multicolumn{3}{c}{{\color[HTML]{000000} BUSI}} &
  {\color[HTML]{000000} } &
  \multicolumn{3}{c}{{\color[HTML]{000000} BUSZS}} \\ \cmidrule(lr){2-4} \cmidrule(lr){6-8} \cmidrule(lr){11-13} \cmidrule(l){15-17} 
{\color[HTML]{000000} } &
  {\color[HTML]{000000} 4} &
  {\color[HTML]{000000} 8} &
  {\color[HTML]{000000} 16} &
  {\color[HTML]{000000} } &
  {\color[HTML]{000000} 4} &
  {\color[HTML]{000000} 8} &
  {\color[HTML]{000000} 16} &
  {\color[HTML]{000000} } &
  {\color[HTML]{000000} } &
  {\color[HTML]{000000} 4} &
  {\color[HTML]{000000} 8} &
  {\color[HTML]{000000} 16} &
  {\color[HTML]{000000} } &
  {\color[HTML]{000000} 4} &
  {\color[HTML]{000000} 8} &
  {\color[HTML]{000000} 16} \\ \cmidrule(r){1-8} \cmidrule(l){10-17} 
{\color[HTML]{000000} WinCLIP\cite{jeong2023winclip}} &
  \multicolumn{3}{c}{{\color[HTML]{000000} 79.7}} &
  {\color[HTML]{000000} } &
  \multicolumn{3}{c}{{\color[HTML]{000000} 59.3}} &
  {\color[HTML]{000000} } &
  {\color[HTML]{000000} } &
  {\color[HTML]{000000} -} &
  {\color[HTML]{000000} -} &
  {\color[HTML]{000000} -} &
  {\color[HTML]{000000} } &
  {\color[HTML]{000000} -} &
  {\color[HTML]{000000} -} &
  {\color[HTML]{000000} -} \\
{\color[HTML]{000000} AdaCLIP\cite{cao2024adaclip}} &
  {\color[HTML]{000000} 54.7} &
  {\color[HTML]{000000} 78.7} &
  {\color[HTML]{000000} 86.5} &
  {\color[HTML]{000000} } &
  {\color[HTML]{000000} 66.9} &
  {\color[HTML]{000000} 83.5} &
  {\color[HTML]{000000} 92.8} &
  {\color[HTML]{000000} } &
  {\color[HTML]{000000} LP++\cite{huang2024lp++}} &
  {\color[HTML]{000000} 56.8} &
  {\color[HTML]{000000} \underline{62.3}} &
  {\color[HTML]{000000} \underline{70.1}} &
  {\color[HTML]{000000} } &
  {\color[HTML]{000000} 54.4} &
  {\color[HTML]{000000} 56.5} &
  {\color[HTML]{000000} \underline{62.2}} \\
{\color[HTML]{000000} AnomalyCLIP\cite{zhou2023anomalyclip}} &
  {\color[HTML]{000000} 62.9} &
  {\color[HTML]{000000} 81.9} &
  {\color[HTML]{000000} 85.6} &
  {\color[HTML]{000000} } &
  {\color[HTML]{000000} 76.6} &
  {\color[HTML]{000000} 87.4} &
  {\color[HTML]{000000} 90.8} &
  {\color[HTML]{000000} } &
  {\color[HTML]{000000} ClipAdapter\cite{gao2024clip}} &
  {\color[HTML]{000000} 51.8} &
  {\color[HTML]{000000} 51.8} &
  {\color[HTML]{000000} 52.0} &
  {\color[HTML]{000000} } &
  {\color[HTML]{000000} 50.7} &
  {\color[HTML]{000000} 50.5} &
  {\color[HTML]{000000} 51.0} \\
{\color[HTML]{000000} VCP-CLIP\cite{qu2024vcp}} &
  {\color[HTML]{000000} \underline{84.8}} &
  {\color[HTML]{000000} \underline{85.7}} &
  {\color[HTML]{000000} 90.0} &
  {\color[HTML]{000000} } &
  {\color[HTML]{000000} 77.3} &
  {\color[HTML]{000000} 86.8} &
  {\color[HTML]{000000} 89.6} &
  {\color[HTML]{000000} } &
  {\color[HTML]{000000} TipAdapter\cite{zhang2021tip}} &
  {\color[HTML]{000000} \underline{58.9}} &
  {\color[HTML]{000000} 61.4} &
  {\color[HTML]{000000} 68.8} &
  {\color[HTML]{000000} } &
  {\color[HTML]{000000} \underline{59.0}} &
  {\color[HTML]{000000} 57.2} &
  {\color[HTML]{000000} 58.7} \\
{\color[HTML]{000000} MVFA\cite{zhang2024mediclipadaptingclipfewshot}} &
  {\color[HTML]{000000} 78.4} &
  {\color[HTML]{000000} 84.0} &
  {\color[HTML]{000000} \underline{90.6}} &
  {\color[HTML]{000000} } &
  {\color[HTML]{000000} \underline{83.3}} &
  {\color[HTML]{000000} \underline{88.9}} &
  {\color[HTML]{000000} \underline{94.8}} &
  {\color[HTML]{000000} } &
  {\color[HTML]{000000} COOP\cite{zhou2022learning}} &
  {\color[HTML]{000000} 58.8} &
  {\color[HTML]{000000} 61.8} &
  {\color[HTML]{000000} 67.7} &
  {\color[HTML]{000000} } &
  {\color[HTML]{000000} 57.9} &
  {\color[HTML]{000000} \underline{58.9}} &
  {\color[HTML]{000000} 62.1} \\
{\color[HTML]{000000} Ours} &
  {\color[HTML]{000000} \textbf{87.0}} &
  {\color[HTML]{000000} \textbf{90.7}} &
  {\color[HTML]{000000} \textbf{91.4}} &
  {\color[HTML]{000000} } &
  {\color[HTML]{000000} \textbf{95.9}} &
  {\color[HTML]{000000} \textbf{98.2}} &
  {\color[HTML]{000000} \textbf{98.2}} &
  {\color[HTML]{000000} } &
  {\color[HTML]{000000} Ours} &
  {\color[HTML]{000000} \textbf{68.6}} &
  {\color[HTML]{000000} \textbf{72.0}} &
  {\color[HTML]{000000} \textbf{73.8}} &
  {\color[HTML]{000000} } &
  {\color[HTML]{000000} \textbf{78.9}} &
  {\color[HTML]{000000} \textbf{82.8}} &
  {\color[HTML]{000000} \textbf{84.1}} \\ \bottomrule
\end{tabular}%
}
\label{tab:cls_results}
\end{table}

\textbf{Anomaly Detection and Classification}
We conduct comprehensive evaluations on two distinct few-shot classification tasks: anomaly detection (binary) and anomaly classification (multi-class). For anomaly detection, we formulate the task as a binary classification problem and evaluate against leading CLIP-based AD approaches \cite{jeong2023winclip, zhou2023anomalyclip, cao2024adaclip, qu2024vcp, huang2024adapting}. Tab.\ref{tab:cls_results} (left) demonstrate UltraAD's superior performance across all settings, particularly in low-shot scenarios where it achieves a substantial 10\% gain over baselines in the challenging 4-shot setting. In addition to anomaly detection, we address the more complex task of anomaly classification, which involves fine-grained classification of anomaly types under limited data constraints. Comparing against state-of-the-art CLIP-based few-shot learning techniques including advanced prompt tuning and adapter-based methods\cite{zhou2022learning,huang2024lp++,gao2024clip,zhang2021tip}, our results in Tab.\ref{tab:cls_results} (right) shows UltraAD's significant advantages, especially on the complex BUSZS dataset, achieving 80\% score on most settings. We attribute this superior performance to our novel integration of segmentation-guided learning. Additionally, we employ a mask-guided post-processing technique, commonly used in recent anomaly detection methods \cite{jeong2023winclip, zhou2023anomalyclip}, which utilizes segmentation masks to refine the separation between normal and abnormal classes. This approach further enhances overall performance, specifically using the formulation $y_{pred} = \frac{1}{2}(\max(\hat{Y}) + \hat{y})$.
\begin{figure}[h]
\centering
\includegraphics[width=\textwidth]{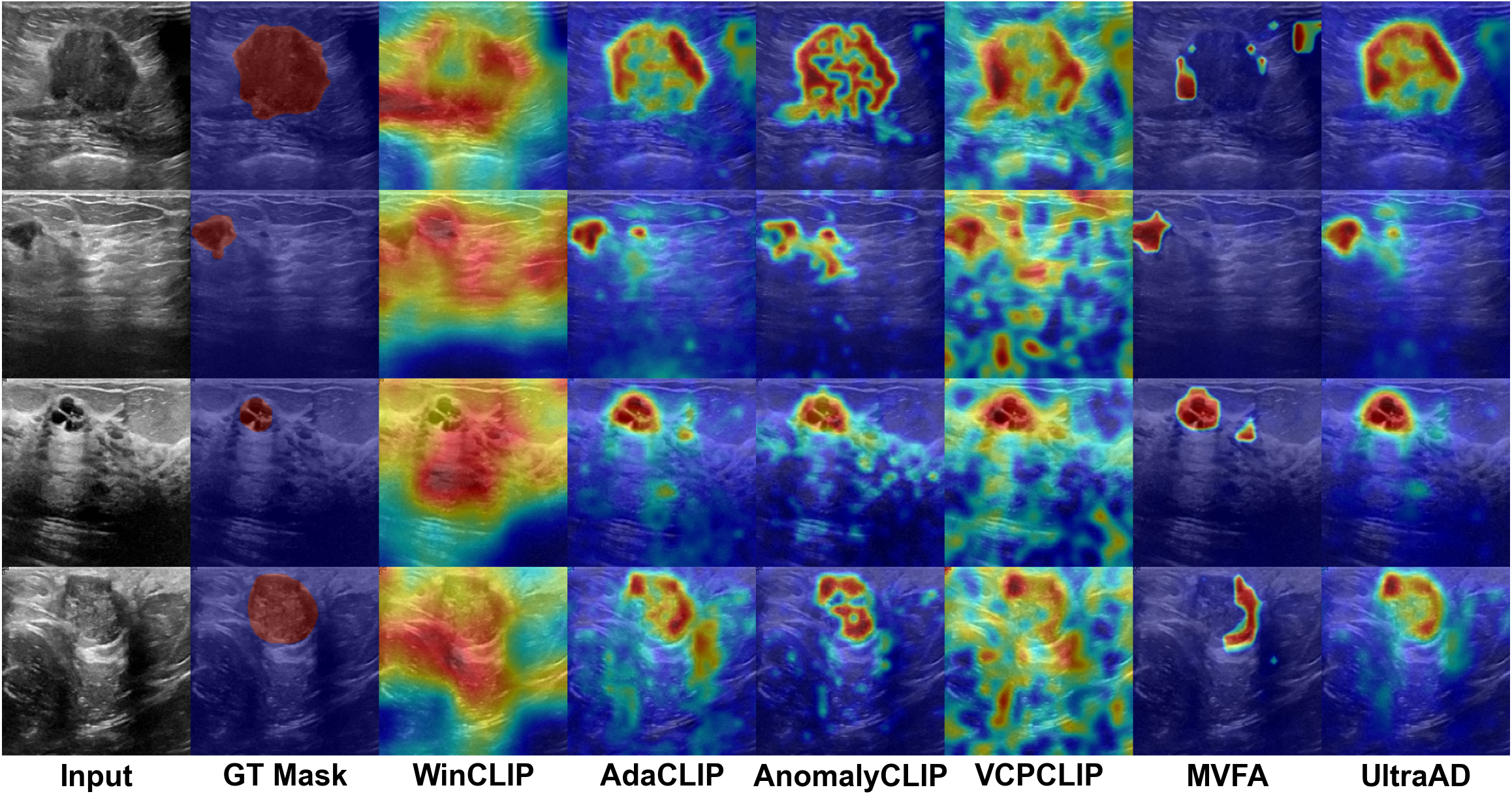}
\caption{\textbf{Visualization of anomaly detection maps from multiple baseline methods.} UltraAD achieves the most precise localization on unseen ultrasound images, effectively identifying anomalies as cohesive regions rather than fragments.}
\label{fig:quantitative_result}
\end{figure}

\begin{table}[]
\centering

\caption{Comparison of anomaly localization performance across benchmarks, evaluated using pixel-level AUROC and AUPRC, shown as (AUROC, AUPRC). Best values are \textbf{bold}, and seconds are \underline{underlined}.}
\resizebox{\columnwidth}{!}{%
\begin{tabular}{@{}cccccccc@{}}
\toprule
\textbf{} &
  \multicolumn{3}{c}{BUSI} &
   &
  \multicolumn{3}{c}{BUSZS} \\ \cmidrule(lr){2-4} \cmidrule(l){6-8} 
Method &
  4 &
  8 &
  16 &
   &
  4 &
  8 &
  16 \\ \midrule
WinCLIP\cite{jeong2023winclip} &
  \multicolumn{3}{c}{(79.7, 5.4)} &
   &
  \multicolumn{3}{c}{{\color[HTML]{333333} (62.1, 8.4)}} \\
AdaCLIP\cite{cao2024adaclip} &
  (79.4, 32.7) &
  (87.5, 51.5) &
  (86.8, 52.4) &
   &
  {(85.9, 42.1)} &
  {(94.3, 60.8)} &
  {(95.4, 70.5)} \\
AnomalyCLIP\cite{zhou2023anomalyclip} &
  (\underline{87.8}, \textbf{50.2}) &
  (\underline{90.0}, \underline{54.8}) &
  (\underline{90.6}, \underline{56.0}) &
   &
  { (\textbf{94.1}, \textbf{64.4})} &
  {(\underline{96.1}, \underline{68.9})} &
  { (\underline{96.6}, \underline{70.9})} \\
VCP-CLIP\cite{qu2024vcp} &
  (75.0, 27.6) &
  (79.0, 34.4) &
  (80.4, 38.4) &
   &
  {(75.3, 21.3)} &
  {(81.7, 32.0)} &
  { (83.5, 36.2)} \\
MVFA\cite{huang2024adapting} &
  (69.9, 16.1) &
  (66.2, 15.1) &
  (69.5, 16.2) &
   &
  {(87.9, 42.4)} &
  { (92.4, 65.0)} &
  {(93.5, 68.9)} \\
Ours &
  (\textbf{88.3}, \underline{48.7}) &
  (\textbf{91.5}, \textbf{58.9}) &
  (\textbf{91.8}, \textbf{58.7}) &
   &
  { (\underline{93.9}, \underline{61.1})} &
  {(\textbf{96.5}, \textbf{70.5})} &
  {(\textbf{96.7}, \textbf{73.3})} \\ \bottomrule
\end{tabular}%
}
\label{tab:seg}
\end{table}

\noindent
\textbf{Anomaly Localization} 
Tab.~\ref{tab:seg} presents a comprehensive analysis of our method against state-of-the-art baseline models for anomaly localization. Our proposed UltraAD demonstrates superior performance compared to most existing methods.
AnomalyCLIP\cite{zhou2023anomalyclip} demonstrates competitive performance attributed to its V-V attention mechanism, which significantly enhances local feature perception. Nevertheless, AnomalyCLIP exhibits limitations in image level anomaly detection show in Tab.\ref{tab:cls_results}. Fig.\ref{fig:quantitative_result} presents the quantitative results of localization. WinCLIP\cite{jeong2023winclip} and VCPCLIP\cite{qu2024vcp} failed to produce satisfactory maps, while MVFA\cite{huang2024adapting} and AnomalyCLIP\cite{zhou2023anomalyclip} struggled to generate continuous regions as anomaly areas. AdaCLIP\cite{cao2024adaclip} included some normal regions in its predictions. Our method detects anomaly locations and predicts them as continuous regions, which can be further used for precise segmentation.

\subsection{Ablation Study}
\label{sec:ab}
We evaluate UltraAD modules via an ablation study in a 4-shot (one-seed) setting on BUSZS Dataset. The results are reported using the notation (I-AUROC/P-AUROC). Our final method achieves scores of (82.3/93.4). Specifically, we examine the impact of employing US-specific terminologies by using US-unaware prompts to generate $\textbf{P}_t$, resulting in scores of (72.3/88.6). To verify the functionality of Memory-Boosted Few-Shot Adaptation, we remove classification and focus on segmentation loss, reducing localization performance and yielding (-/84.3) without joint learning. To verify the Patch-Wise Image-Language Fusion, we omit the Patch-Wise Image-Language Fusion module, instead directly utilizing the linearly projected patch tokens $f'_p$ for anomaly localization, resulting in (71.2/90.3). 


\section{Conclusion}
We introduce UltraAD, an innovative framework that enables simultaneous abnormal localization and fine-grained classification in US imaging through few-shot adaptation of the CLIP model. Few-shot adaptation was applied to a breast dataset and validated on two unseen datasets with different machines, patients, and probes, demonstrating robustness in real-world scenarios. An ablation study further validates the effectiveness of the memory-boosted few-shot adaptation and PIF module. The promising results on breast US images highlight the potential for developing a generalized model for US imaging, with future extensions to diverse anatomies using corresponding datasets.


\begin{credits}
\subsubsection{\ackname} This work was supported in part by the Multi-Scale Medical Robotics Center, AIR@InnoHK, Hong Kong; and in part by the SINO-German Mobility Project under Grant M0221.

\subsubsection{\discintname}
The authors have no competing interests to declare that are
relevant to the content of this article.
\end{credits}

\bibliographystyle{splncs04}
\bibliography{reference}
\end{document}